\pdfoutput=1
\documentclass[11pt]{article}

\usepackage[final]{acl}

\usepackage{times}
\usepackage{latexsym}

\usepackage[T1]{fontenc}
\usepackage[utf8]{inputenc}

\usepackage{microtype}

\usepackage{inconsolata}

\usepackage{graphicx}

\usepackage{tabularx}
\usepackage{hyperref}       % hyperlinks
\usepackage{url}            % simple URL typesetting
\usepackage{booktabs}       % professional-quality tables
\usepackage{nicefrac}       % compact symbols for 1/2, etc.
\usepackage{microtype}      % microtypography
\usepackage{xcolor}         % colors
\usepackage{lipsum}
\usepackage{multirow}
\usepackage{stmaryrd}
\usepackage{amsmath}
\usepackage{multirow}
\usepackage{bbm}
\usepackage{graphicx}
\usepackage{amssymb}
\usepackage{adjustbox}
\usepackage{xspace}
\usepackage{algpseudocode}
\usepackage{algorithm}
\usepackage{graphicx}
\usepackage{caption}
\usepackage{subcaption}
\usepackage{color, colortbl}
\usepackage{tablefootnote}
\usepackage{enumitem}  % More control
\usepackage{lipsum}
\usepackage{xspace}
\usepackage{gensymb}
\usepackage{textcomp}
\usepackage[normalem]{ulem}  %for strikeout
\usepackage{CJKutf8}  %for CJK 

\setlist{topsep=1pt,itemsep=1pt,partopsep=1pt, parsep=1pt}

\newcommand\datasetname{StingrayBench}

\title{Thank You, Stingray: Multilingual Large Language Models \\ Can Not (Yet) Disambiguate Cross-Lingual Word Sense}

\author {
    \textbf{Samuel Cahyawijaya}$^{*1,3,7}$\quad
    \textbf{Ruochen Zhang}$^{*2,3}$\quad
    \textbf{Holy Lovenia}$^{*3,7}$\quad
    \\
    \textbf{Jan Christian Blaise Cruz}$^{*4,3}$\quad
    \textbf{Elisa Gilbert}$^{5}$\quad
    \textbf{Hiroki Nomoto}$^{*6,3}$\quad
    \textbf{Alham Fikri Aji\thanks{\;\,Equal contribution. }}$^{4,3,7}$\quad
    \
    \\
    $^1$Cohere \;
    $^2$Brown University \;
    $^3$SeaCrowd \; 
    $^4$MBZUAI \;\\
    $^5$University of T\"ubingen\;
    $^6$TUFS \; 
    $^7$IndoNLP \; \\
    \texttt{samuelcahyawijaya@cohere.com; ruochen\_zhang@brown.edu;} \\ 
    \texttt{holylovenia@gmail.com; jan.cruz@mbzuai.ac.ae; nomoto@tufs.ac.jp} \\
    \texttt{elisa.gilbert@student.uni-tuebingen.de;
    alham.fikri@mbzuai.ac.ae} \\ 
}

\begin{document}
\maketitle
\begin{abstract}

Multilingual large language models (LLMs) have gained prominence, but concerns arise regarding their reliability beyond English. This study addresses the gap in cross-lingual semantic evaluation by introducing a novel benchmark for cross-lingual sense disambiguation, \datasetname{}\footnote{To ensure reproducibility, we release our benchmark at \url{https://huggingface.co/datasets/StingrayBench/StingrayBench} under the CC-BY-SA 4.0 license and evaluation suite at \url{https://github.com/SamuelCahyawijaya/stingraybench} under the Apache-2.0 license.}. In this paper, we demonstrate using false friends---words that are orthographically similar but have completely different meanings in two languages--- as a possible approach to pinpoint the limitation of cross-lingual sense disambiguation in LLMs. We collect false friends in four language pairs, namely Indonesian-Malay, Indonesian-Tagalog, Chinese-Japanese, and English-German; and challenge LLMs to distinguish the use of them in context. In our analysis of various models, we observe they tend to be biased toward higher-resource languages. We also propose new metrics for quantifying the cross-lingual sense bias and comprehension based on our benchmark. Our work contributes to developing more diverse and inclusive language modeling, promoting fairer access for the wider multilingual community. 

% https://arxiv.org/html/2404.00929v1
% https://arxiv.org/abs/2310.10310
% https://aclanthology.org/2023.ijcnlp-short.9/

\end{abstract}

\section{Introduction}

% Large language models (LLMs) have become integral tools in a variety of applications, from chatbots and writing assistants to content creation. While these models have advanced our capabilities, there are growing concerns about the presence of bias in their outputs. Bias in LLMs can manifest in various forms, and ensuring fairness and accuracy in their responses is crucial. In particular, the issue of language bias in multilingual LLMs has come under increasing scrutiny. Multilingual LLMs are designed to understand and generate text in multiple languages, but they may exhibit biases towards certain languages or language families. A recent study by \citet{Nomoto23-chatgpt} shed light on a specific type of language bias, termed "language selection bias." This bias occurs when a multilingual LLM generates responses in a different language from the one it was prompted with. For example, the model might reply in Indonesian when prompted in Malay, leading to a significant misrepresentation of the user's intent. 

Multilingual large language models (LLMs) have become integral tools in a variety of tasks and languages~\cite{bang-etal-2023-multitask,yong-etal-2023-prompting,zhang-etal-2023-multilingual,lovenia2024seacrowdmultilingualmultimodaldata,cahyawijaya2024llmeveryone,cahyawijaya-etal-2024-cendol}. 
% On the foundation of these amazing LLMs lies token embeddings which are the unitary building block of most of the existing LLMs. 
While these LLMs have remarkable capabilities, there are growing concerns about the reliability of their responses especially in languages other than English. Most evaluations address cross-lingual generalization in LLMs by assessing their ability on the set of downstream tasks as the one used in English~\cite{cahyawijaya-etal-2021-indonlg,adelani-etal-2023-masakhanews,kabra-etal-2023-multi,zhang2023miracl,adelani-etal-2024-sib,cahyawijaya-etal-2024-cendol,zhang-eickhoff-2024-crocosum}, many even directly translated from the source corpora~\cite{hu2020xtreme,cahyawijaya-etal-2021-indonlg,winata-etal-2023-nusax,cahyawijaya-etal-2023-nusawrites,bandarkar-etal-2024-belebele,singh-etal-2024-aya}. These evaluations reflect the cross-lingual generalization in the downstream application level, but fail to capture the basic understanding of semantic meaning across different languages. This lack of semantic understanding further extends to the unexplained bias of multilingual LLMs towards certain languages or language families which causes the LLMs to respond in their preferred languages, leading to a significant misrepresentation of users' intent~\cite{Nomoto23-chatgpt,nomoto2024masalah}.

\begin{figure}
    \centering
    \includegraphics[width=\linewidth]{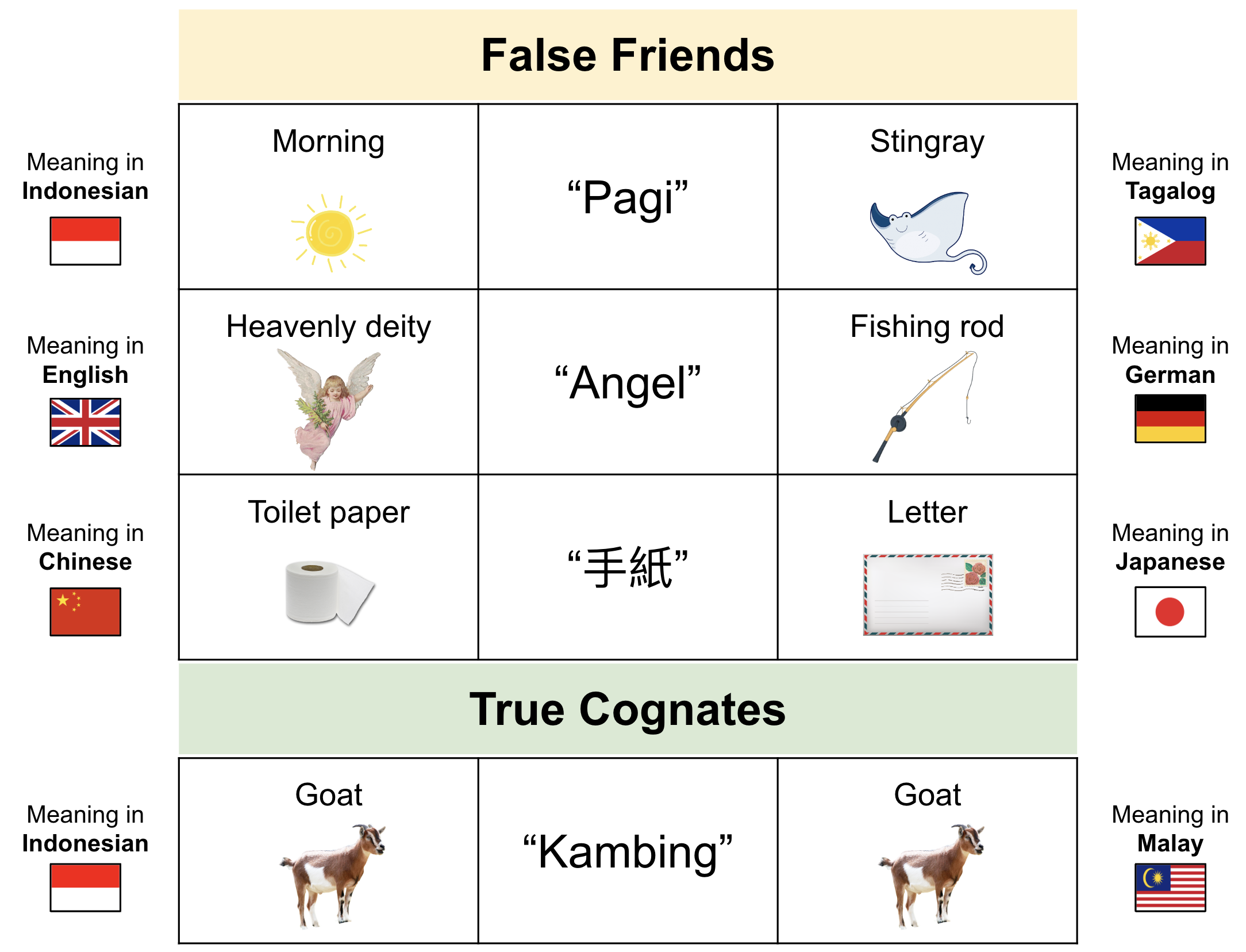}
    \caption{Our work explores two linguistic phenomena known as \textbf{false friend} and \textbf{true cognate}, and highlights the limitation of LLMs on understanding cognate indicating the pitfall on cross-lingual disambiguation.}
    \vspace{-8pt}
    \label{fig:stingray_ff_vs_tc}
\end{figure}

% Our work aims to build upon and extend these findings by generalizing the understanding of language selection bias and exploring its underlying causes. We focus on the concept of "cognate"\footnote{We use the term ``cognate'' to refer only to non-false friend cognates.  Some false friends are in fact cognates.  Thus, \textit{bandar} meaning `port' in Indonesian and `city' in Malay originates from the same language but came to convey different meanings as a result of language change.  Such expressions are not called ``cognates'' here.}
%  and "false friends," which are words or phrases that sound similar in two different languages but have distinct meanings\footnote{Our title "Thank You, Stingray" is a playful reference to a false friend phrase "Selamat Pagi", which means "Good morning" in Indonesian but means "Thank you, stingray" in Tagalog, which might bring confusion to multilingual LLMs.} and introduce the first benchmark for measuring language selection bias dubbed as \datasetname{}. By analyzing false friends in multiple language pairs, including Indonesian-Malay, Indonesian-Tagalog, Chinese-Japanese, and English-German, we strive to uncover patterns and potential sources of language selection bias. Through a meticulous task formulation, we delve into the internal workings of LLMs to identify factors contributing to this type of bias.

Our work aims to explore the cross-lingual evaluation of semantic meaning in LLMs and understand its underlying causes. We focus on the concept of "false friends", which are words or phrases that sound similar in two languages but have distinct meanings\footnote{Our title "Thank You, Stingray" is a playful reference to a false friend phrase "Selamat Pagi", which means "Good morning" in Indonesian but means "Thank you, stingray" in Tagalog, which might bring confusion to multilingual LLMs.} and "true cognates", which are words or phrases that sound similar in two languages and share the same meaning\footnote{We use the term \textbf{common words} as the umbrella term to refer to both true cognates and false friends.}. We create data instances containing false friends and true cognates as described in Figure~\ref{fig:stingray_ff_vs_tc}. Using these concepts, we construct the first benchmark for measuring cross-lingual semantic understanding in LLMs dubbed as \datasetname{}. By analyzing LLMs performances on \datasetname{} containing multiple language pairs, we assess whether they exhibit language-selection bias through the task of cross-lingual sense disambiguation with new metrics and present future research directions to mitigate these biases. 
% Through meticulous task formulation, we delve into the internal workings of LLMs to identify factors contributing to this type of bias.
Our contributions and the significance of this work can be summarized as follows:
\begin{itemize}
    \item We propose \textbf{\datasetname{}}, the first benchmark for measuring the cross-lingual sense disambiguation in LLMs covering four distinct language pairs, i.e., Indonesian-Malay (\textsc{id}-\textsc{ms}), Indonesian-Tagalog (\textsc{id}-\textsc{tl}), Chinese-Japanese (\textsc{zh}-\textsc{ja}), and English-German (\textsc{en}-\textsc{de}).
    \item We introduce a method to measure cross-lingual sense comprehension and bias in LLMs by introducing \textbf{stringray plot} and two evaluation metrics for measuring cross-lingual sense understanding, i.e., \textbf{cognate bias} and \textbf{cognate comprehension score}.
    \item We showcase the generalization of the cognate bias phenomena to multiple multilingual LLMs in diverse language pairs, demonstrating its broader impact and severity in existing multilingual LLMs.
    % \item We highlight the potential source of the bias in multilingual LLMs, bringing insights for mitigating the language selection bias.
\end{itemize}

\section{Related Works}

\subsection{Cognates and False Friends in NLP} 
Homologous words that show systematic sound correspondences indicating common 
ancestry are known as cognates~\cite{atkinson2013descent}. For example, \textit{baru} in Malay and \textit{bago} in Tagalog are cognates based on the systematic \textit{r-g} sound correspondence, both meaning `new'.  However, cognates do not necessarily have the same meaning, as is the case with \textit{bibir} meaning `lip' in Malay and \textit{bibig} `mouth' in Tagalog, which show the same \textit{r-g} correspondence. The study of cognacy contributes to understanding the historical lineage of languages and the reconstruction of proto-languages~\cite{campbell2013historical}. 

Many recent works focus on the identification of cognates in genetically related languages~\cite{batsuren-etal-2019-cognet, Batsuren2021ALA, Bafna2022CombiningNS, Dinu2023RoBoCoPAC, akavarapu-bhattacharya-2024-automated}. 
One of the factors that make cognate identification non-trivial is the presence of false friends (or false cognates). False friends are words that are orthographically or phonetically similar but do not share the same meaning~\cite{allan2009concise}.  While many false friends are indeed cognates, some are not true cognates and can be mistaken for cognates. For example, an Indonesian-Malay false friend \textit{polisi} `police (Indonesian), policy (Malay)' traces back to different ancestor languages, i.e.\ Dutch (\textit{politie} `police') and English (\textit{policy}). Besides posing a challenge to cognate identification, false friends also constitute a major obstacle for translators, language learners and especially machine translation systems. Studying false friends is not easy because it requires bilingual proficiency and as a result, false friends have received little attention. Current studies on false friends focus on their collection and identification \citet{Ljubesic2013IdentifyingFF,castro-etal-2018-high, uban2020automatically}. In our paper, we deal with false friends in multiple language pairs and use them as a tool to understand the proficiency of multilingual large language models. 

\begin{figure*}
    \centering
    \includegraphics[width=\linewidth]{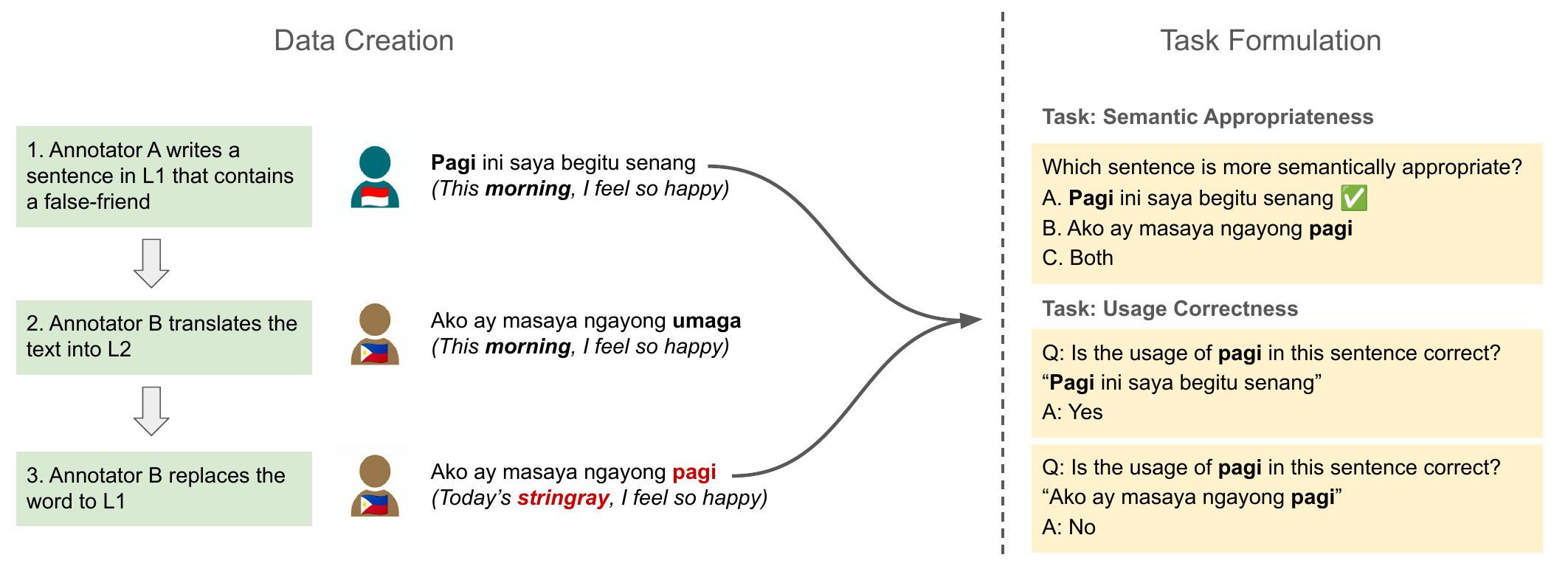}
    \caption{Annotation and data formulation pipeline of \datasetname{}. Our annotation consists of a 3-step process that requires two annotators, one for each language of the language pair. In addition, we provide the English translation of the correct sentence for better accessibility to \datasetname{}.}
    \vspace{-8pt}
    \label{fig:stingray_bench_pipeline}
\end{figure*}

\subsection{Word Sense Disambiguation} 
Word Sense Disambiguation (WSD) task aims to determine the correct meaning of a polysemous word in a given context~\cite{bevilacqua2021recent, bevilacqua2020breaking, blevins2020moving}
The task has also been extended to a multilingual setting~\cite{navigli2013semeval, pasini2021knowledge, pasini2021xl, su2022multilingual}, leveraging multilingual lexical knowledge bases~\cite{navigli2012babelnet, bond2013linking}. It has been a challenging task in NLP since its early recognition by~\citet{j:weaver} and remains critical in recent works that investigate ``the curse of multilinguality''~\cite{conneau2019unsupervised, berend2023combating} and universal representations across languages~\cite{wu2019emerging, wendler2024llamas, ferrando2024similarity, Zhang2024TheSB}. Built upon previous studies, our work explores the ability of multilingual LLMs to disambiguate word sense across languages. The closest related task is the Word-in-Context (WiC) task~\cite{pilehvar2018wic, raganato2020xl}, where they ask to classify the word usage given two distinct contexts in the same language. In contrast, we embed the word in parallel contexts across different languages, with only one sentence being semantically correct. By evaluating whether LLMs align the word's meaning with the appropriate language-specific context, we can assess their multilingual capabilities and detect potential language selection bias.

\section{\datasetname{}}

\subsection{Dataset Construction}

To construct \datasetname{}, native speakers of Chinese, English, German, Indonesian, Japanese, Malay, and Tagalog are asked to list down common words that existed between the following language pairs together with their meanings: English-German, Indonesian-Malay, Indonesian-Tagalog, and Chinese-Japanese. For this, the following resources dealing with false friends are consulted: Wiktionary's lists of false friends,\footnote{\url{https://en.wiktionary.org/wiki/Category:False_cognates_and_false_friends}} \textit{Kamus Komunikatif Nusantara: Indonesia-Malaysia, Malaysia-Indonesia} \citep{MohdSharifudinAlMudra2015}, and \textit{Kamus Kata: Bahasa Melayu Malaysia-Bahasa Indonesia} \citep{Rusdi2016}. Some words in these resources are rejected as they turned out not to be false friends after scrutiny.  Moreover, it is not always easy to find common words with identical spellings and characters, except for Indonesian-Malay. Therefore, we allow the use of words differing in capitalization (e.g.\ \textit{arm}-\textit{Arm}) in English-German, the use of words with one edit distance (e.g.\ \textit{aku}-\textit{ako}) in Indonesian-Tagalog, and the use of words with different characters developed from the same origin (e.g.\ \begin{CJK*}{UTF8}{gbsn}图书馆\end{CJK*}-\begin{CJK*}{UTF8}{ipxm}図書館\end{CJK*}) in Chinese-Japanese. These common words are then segregated into false friends (same word with different meanings) and true cognates (same word with the same meaning).

For each word, annotators would construct a sentence that uses that word in their native language. An English translation would then be written by the annotator. The annotator of the other language in the pair would then construct a sentence in their native language that follows the meaning of the sentence in the first language and/or that of the English translation. For sentences involving a false friend, an accurate translation would not employ the target false friend word but a different word that expresses the intended meaning in the language. Hence, an additional step of replacing the latter word with the target false friend word is required, which produces semantically odd sentences.

For example, in Indonesian-Tagalog, for the word \textit{pagi} meaning `morning' in Indonesian and `stingray' in Tagalog. The Indonesian annotator would construct \textit{Pagi ini saya begitu senang} as the sentence. The English translation would be \textit{This morning, I feel so happy}. The Tagalog annotator would then translate it as \textit{Ako ay masaya ngayong \textbf{umaga}} and replace the word meaning `morning', i.e., \textit{umaga}, by the target false friend word \textit{pagi} to produce \textit{Ako ay masaya ngayong \textbf{pagi}}, which means `Today's stingray, I feel happy' and is semantically odd in Tagalog. 

%In the case of true cognates, both annotators would agree on an English sentence to be constructed first, then that sentence is translated to their native languages. For example, the annotators of Indonesian-Malay would agree on "That apple has many caterpillars." as the English sentence for "Ulat" which means "Catterpillar" in both languages. The Indonesian annotator would write "Apel itu banyak ulat." and the Malaysian annotator would write "Epal itu ada banyak ulat.", respectively. \hn{[\%Do we really need this paragraph?  The next paragraph would be easier to understand without this intervening paragraph.]}

In most cases, each false friend will have two entries in the final dataset corresponding to one correct and one incorrect usage of that word. However, in some cases, a false friend has only one entry. This happens for partial cognates: when the word shares the same meaning in two languages but has an additional meaning in one language but is absent in the other. For example, \textit{pelatih} means `trainer' in both Indonesian and Malay, but it has another meaning in Malay, but not in Indonesian, i.e.\ `trainee'. Each true cognate will only have one entry as both native language sentences translate to each other correctly. The detailed annotation guideline is provided in Appendix~\ref{sec:annotation-guideline}. The statistics of the \datasetname{} are described in Table \ref{tab:dataset_size}.

\begin{table}[!t]
\small
\centering
\resizebox{0.95\columnwidth}{!}{%
% \begin{tabularx}{\columnwidth}{p{1.2cm}p{1cm}p{1.8cm}p{1.8cm}}
\begin{tabular}{cccc}
    \toprule
    \multicolumn{1}{c}{\textbf{Subset}} & \textbf{\#True Cognate} & \textbf{\#False Friend} & \textbf{\#Total} \\
    \midrule
    \textsc{en-de} & 98 & 98 & 196 \\
    \textsc{id-ms} & 52 & 134 & 186 \\
    \textsc{id-tl} & 58 & 100 & 158 \\
    \textsc{zh-ja} & 51 & 114 & 165 \\
    \midrule
    \textbf{Total} & \textbf{259} & \textbf{446} & \textbf{705} \\
    \bottomrule
\end{tabular}%
}
\caption{Statistics of our \datasetname{}.}
\label{tab:dataset_size}
% \vspace{-10pt}
\end{table}

\subsection{Task Formulation}
\label{sec:task-formulation}
Using the sentences collected above, we propose two task formulations of different semantic granularities as follows. Notice that we prompt the model in English as it is language-neutral for most of the language pairs except for the English-German case.

\paragraph{Semantic Appropriateness} Given the data construction as described above, we want to test the models' competence for sentence comprehension. In this task, we prompt the model with: \textit{Which sentence is more semantically appropriate?}. The first two options are the two sentences for the language pair respectively. A third option, that both sentences are appropriate, is also included. It is the correct option for the true cognates scenario but also serves as a confounding option for the false cognates subset. We provide the example of the prompt and the target completion for the semantic appropriateness task in Figure~\ref{fig:stingray_bench_pipeline}.

\paragraph{Usage Correction} In this task formulation, we emphasize the usage of the specific cognate words by prompting the models with: \textit{Is the usage of [WORD] in this sentence correct? [SENTENCE]}. We expect this task to be simpler as 1) the options are binary with no confounding options; and 2) specific cognates are mentioned in the prompt which potentially serves as a task hint. We provide the example of the prompt and the target completion for the usage correction task in Figure~\ref{fig:stingray_bench_pipeline}.

\subsection{Measure of Cognate Understanding Ability in LLMs}

\begin{figure}[!t]
  \centering
  \includegraphics[trim={0, 5mm, 0, 0}, clip, width=\linewidth]{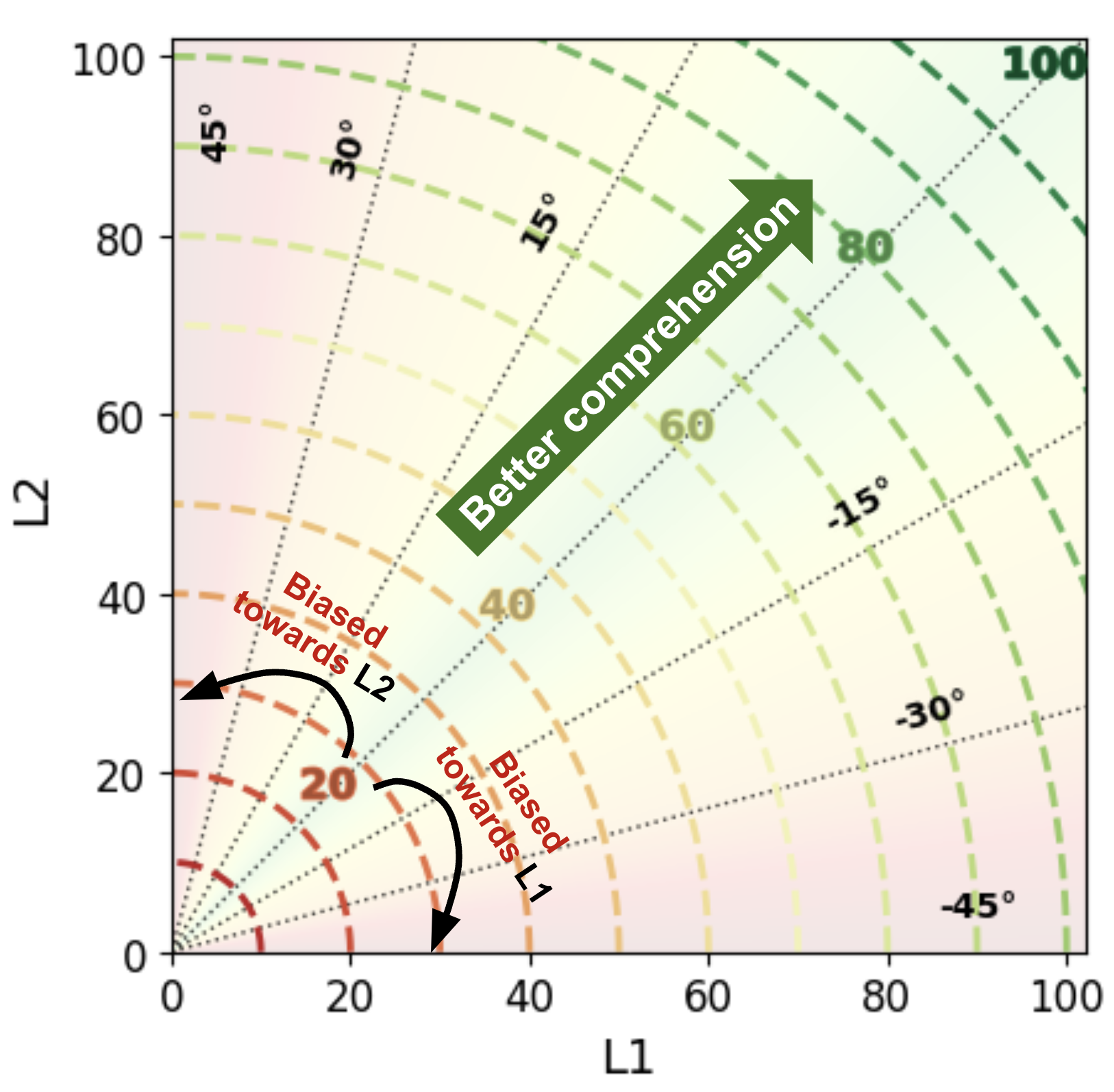}
  % \caption{Stingray  Plot}
  % \label{fig:cognate-pair-plot}
  \caption{Stingray plot is a 2D scatter plot where the X-axis and Y-axis represent the model performance on StingrayBench across each language.  The \textbf{cognate bias score} towards a particular language is measured based on the angular distance of the data point (e.g. the model is unbiased if it has equally good performance for either language). The \textbf{cognate comprehension score} is measured based on the point's magnitude. 
   % which indicates whether a model have a certain language bias on understanding cognate within a certain language pair,
  % which indicates how good a model performs on differentiating cognate and false friend within a certain language pair.
  }
  % \caption{SEACrowd covers 980 languages in Southeast Asia. There are $\sim$300 languages uncovered Southeast Asian languages remain after SEACrowd.}
  \label{fig:cognate_method}
  \vspace{-10pt}
\end{figure}

We define cognate understanding as the ability of LLMs to be able to correctly comprehend the semantic meaning of a cognate for both "true cognate" and "false friend". This is done through probing LLMs with questions that ensure the understanding of LLMs to the semantic nuances of the cognate, which can be either "true cognate" or "false friend", in the context of the relevant language pairs as described in Section~\ref{sec:task-formulation}. Using the tasks in \datasetname{}, we measure the per-language accuracy of LLMs on each task and conduct further analysis as described below.

\paragraph{Stingray Plot}

To measure cognates understanding ability of LLMs on a certain language pair $<L_1, L_2>$, we need to take into account the cognate understanding quality on both $L_1$ and $L_2$. 
% In this case, we needs to consider at least two dimensions one for representing the cognate quality in $L_a$ and the other to represent the cognate quality in $L_b$. 
To do so, we derive our analysis based on a 2-dimensional vector space and introduce the Stingray plot. As shown in Figure~\ref{fig:cognate_method}, the Stingray plot presents two different contours: (1) a U-shaped angular contour with a minimum value of 0 at either 0$\degree$ and 90$\degree$ angle and a maximum value of 1 at 45$\degree$ angle; and (2) the radial contour with a minimum value of 0 at the bottom left corner and a maximum value of 100 at the top right corner. Using this characteristic of the Stingray plot, we develop two metrics for measuring cognate understanding, i.e., \textbf{cognate bias} and \textbf{cognate comprehension}.

\begin{table}[!t]
    \centering
    \resizebox{\linewidth}{!}{
        \begin{tabular}{lcc}
        \toprule
        \multicolumn{1}{c}{\textbf{Model Name}} & \multicolumn{1}{c}{\textbf{Model Size}} & \multicolumn{1}{c}{\textbf{Supported Lang.}} \\
        \bottomrule
        \multicolumn{3}{c}{\cellcolor{lightgray}\textit{Monolingual / Bilingual LLMs}} \\
        ChatGLM2 & 6B & en, zh \\
        Yi-1.5 & 9B, 34B & en, zh \\
        Phi-3 & 3.8B, 7B, 14B & en \\
        Cendol LLaMA-2 & 7B & id, (en, de)$^*$ \\
        Cendol mT5 & 3.7B & id, (en, de, zh, ja, ms, tl)$^*$ \\  
        \midrule
        \multicolumn{3}{c}{\cellcolor{lightgray}\textit{Multilingual LLMs}} \\
        SeaLLM v3 & 7B & en, zh, id, ms, tl \\
        SEA-LION v2.1 & 8B & en, zh, id, ms, tl \\
        BLOOMZ & 0.6B, 1.1B, 1.7B, 3B, 7B & en, zh, id \\
        mT0 & 0.3B, 0.6B, 1.2B, 3.7B, 13B & en, de, zh, ja, id, ms, tl \\
        Aya-101 & 13B & en, de, zh, ja, id, ms, tl \\
        Aya-23 & 8B, 35B & en, de, zh, ja, id, ms, tl \\
        QWEN-2.5 & 0.5B, 1.5B, 3B, 14B, 32B & en, de, zh, ja \\
        Command-R & 35B & en, de, zh, ja, (id)$^*$ \\
        GPT-4o Mini & - & en, de, zh, ja, (id, ms, tl)$^*$ \\
        Llama-3.1 & 8B, 70B & en, de \\
        Llama-3.2 & 1B, 3B & en, de \\
        \bottomrule
        \end{tabular}
    }
    \caption{List of LLMs incorporated in our experiment. For language codes, we adopt the ISO 639-3 standard. Asterisk ($^*$) denotes that the language is not officially supported or is only included in the pre-training phase.}
    \label{tab:llm}
    % \vspace{-10pt}
\end{table}

\paragraph{Cognate Bias Score}

Given a language pair, an LLM can perform well in identifying cognates in one, but poor in the other. In this case, we can expect that the model has a certain degree of understanding bias in one language. An unbiased LLM should yield similar performance on both languages, while an extremely biased LLM should perform well on one, and close to random estimator for the other. To quantify the \textbf{cognate bias score}, we follow the U-shaped angular contour in the Stingray plot. Specifically, we measure the angular distance between the $<L_1, L_2>$  performance of an LLM with the 45$\degree$ angle. To disambiguate between bias to $L_1$ and $L_2$, we incorporate the sign such that a negative distance indicates a bias towards $L_1$, while a positive distance indicates a bias towards $L_2$. Lastly, we normalize the range of the \textbf{cognate bias score} by linearly scaling from the original range of $[-\frac{\pi}{4}\dots\frac{\pi}{4}]$ to $[-1.0\dots1.0]$.

\paragraph{Cognate Comprehension Score}
Cognate bias shows the understanding of one LLM in a certain language, but it does not reflect the proficiency of LLM in understanding cognates. For instance, when an LLM behaves like a random estimator in both languages, it will yield similar accuracy scores (50\% for binary classification, 33\% for ternary classification, etc) in both languages. In this case, the LLM does not seem to exhibit much cognate bias, but it does not imply that the LLM has an exceptional cognate understanding. To quantify the cognate understanding ability, we introduce the \textbf{cognate comprehension score}. A perfect cognate comprehension score indicates that the LLM is unbiased and performs well in both languages. The cognate comprehension is implemented by simply calculating the magnitude of the $<L_1, L_2>$ vector and normalizing the range into $[0\dots1]$ by dividing the magnitude with $\sqrt{2}$. Note that, when the LLM yields a perfect score on $L_1$ and 0 on $L_2$, the performance only achieves $\sim$70.71\% cognate comprehension score, further improvement from this point will also reduce the bias of the LLM.

\begin{figure*}[!t]
  \centering
  \includegraphics[trim={0, 2mm, 0, 0}, clip, width=0.9\linewidth]{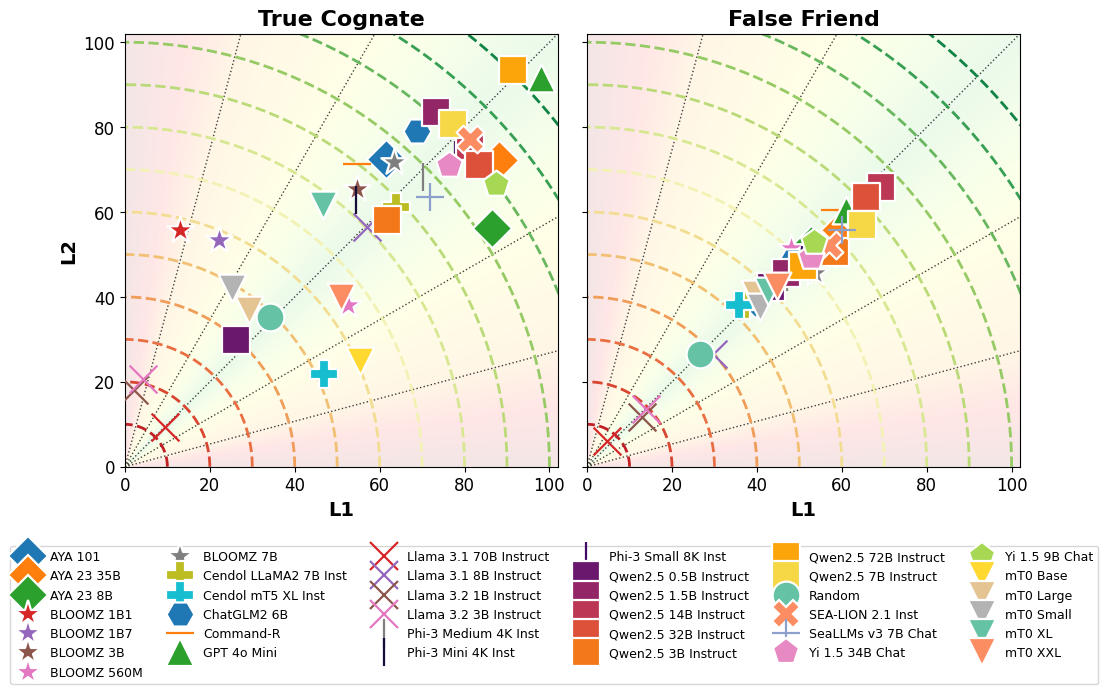}
  \caption{Stingray plot showcasing the performance of each LLM averaged across all language pairs and tasks. There is a different trend between the model performance on the \textbf{(left)} true cognate and \textbf{(right)} false friend subsets. LLMs showcase strong capability on true cognates, but close to random guessing on false friends. This highlights the inability of existing LLMs to disambiguate false friends across different languages.}
  \label{fig:stingray_overall_results}
  % \vspace{-8pt}
\end{figure*}

\section{Experiment Setting}

\subsection{Data Subsets}
We utilize the collected \datasetname{} for our evaluation which covers four language pairs, i.e., English-German, Indonesian-Malay, Indonesian-Tagalog, and Chinese-Japanese. For each language, we split the data into two different subsets based on the phenomenon observed, i.e., true cognate and false friend subsets. As there are only limited amount of data, we aggregate the score from multiple tasks to improve the reliability of the LLMs prediction. The statistics of the \datasetname{} per language pair and per subset are shown in Table~\ref{tab:dataset_size}.

\subsection{Model}

Our evaluation covers a wide variety of LLMs, from monolingual, bilingual, and multilingual LLMs. For multilingual LLMs, we incorporate BLOOMZ~\cite{le2023bloom,muennighoff-etal-2023-crosslingual}, mT0~\cite{muennighoff-etal-2023-crosslingual}, Aya-101~\cite{singh-etal-2024-aya,ustun-etal-2024-aya}, Aya-23~\cite{aryabumi2024aya23openweight}, Qwen-2.5~\cite{qwen2,qwen2.5}, Command-R~\cite{c4ai2024commandr,c4ai2024commandrplus}, and GPT-4o mini~\cite{openai2024gpt4technicalreport}. We also explore LLMs with lower language coverage or specifically adopted for certain languages including Phi-3~\cite{abdin2024phi3technicalreporthighly}, Cendol LLaMA-2~\cite{touvron2023llama2openfoundation,cahyawijaya-etal-2024-cendol},  Cendol mT5~\cite{Xue2020mT5,cahyawijaya-etal-2024-cendol}, SEALLM v3~\cite{nguyen-etal-2024-seallms},  SEA-Lion v2.1~\cite{tjhi-etal-2023-sea}, ChatGLM2~\cite{glm2024chatglmfamilylargelanguage}, and Yi~\cite{ai2024yiopenfoundationmodels}. We exhaustively explore different size variations of each LLM with the scale ranging from 0.3B to 70B parameters to better understand the effect of scaling on the cognate understanding of LLMs. The list of LLMs covered in our study is shown in Table~\ref{tab:llm}.

\subsection{Evaluation \& Inference}
\label{sec:eval}
% For the Chinese-Japanese evaluation, as Chinese characters and Japanese Kanji are encoded in different Unicode, the LLMs will comprehend these characters as two completely different tokens. To see the effect of cognate under the same token embeddings, we standardize these characters by using only the Chinese characters. 
For the inference, we conduct zero-shot prompting by prompting LLMs to answer the given prompt directly using each of the corresponding chat formats supported in each LLM. We perform two different types of inference: (1) likelihood-based inference; and (2) generation-based inference. 

% \begin{figure*}[!t]
%   \centering
%   \includegraphics[trim={0, 0, 0, 0}, clip, width=\linewidth]{images/true_cognate_understanding.png}
%   \caption{Most LLMs understand true cognates in language pairs under study. We report the averaged cognate comprehension scores across the semantic correctness and usage correctness tasks on the true cognate subset.}
%   \label{fig:true_cognate_understanding}
%   % \vspace{-8pt}
% \end{figure*}

% \begin{figure*}[!t]
%   \centering
%   \includegraphics[trim={0, 0, 0, 0}, clip, width=\linewidth]{images/false_friend_understanding.png}
%   \caption{Most LLMs have limited understanding in regards to false friends in language pairs under study. We report the averaged cognate comprehension scores across the semantic correctness and usage correctness tasks on the false friend subset.}
%   \label{fig:false_friend_understanding}
%   \vspace{-8pt}
% \end{figure*}

\paragraph{Likelihood-based}
To perform likelihood-based inference, we follow the zero-shot prompting implementation from prior works~\cite{cahyawijaya-etal-2023-instructalign,cahyawijaya-etal-2023-nusawrites,zhang-etal-2023-multilingual,lovenia2024seacrowdmultilingualmultimodaldata}. For binary classification tasks, we use the label with the highest marginal likelihood given the prompt. For multiple-choice tasks, we provide the choices after the query and take the answer choice label, i.e., \texttt{A}, \texttt{B}, or \texttt{C}, with the highest likelihood. We opt for the likelihood-based for open-source LLMs as we cannot perform this on the API-based LLMs.

\paragraph{Generation-based}
To generalize and ensure the robustness of our results, we also do inference with a generation-based approach. The prompts are shown in \ref{fig:stingray_bench_pipeline} and with an additional sentence that asks LLMs to limit their answers to ``A, B or C'' or ``Yes or No''. To get the final result, we post-process the generated responses: as an example, for the semantic appropriateness task, LLMs sometimes answer ``A and B'' instead of the option ``C''. We test all LLMs listed in Table~\ref{tab:llm}. Nonetheless, we also note that in some LLMs, they often fail to follow the given instructions.

% \paragraph{Chinese-Japanese evaluation} For the Chinese-Japanese evaluation, since Chinese and Japanese characters are encoded in different Unicodes, we use the corresponding characters in each language for the common words. In this case, the resulting sentence looks more natural with the corresponding character sets per language without any abrupt language change in the sentence.

\begin{figure*}[!t]
  \centering
  \begin{subfigure}[t]{\linewidth}
      \includegraphics[trim={0, 6.6em, 0, 0}, clip, width=\linewidth]{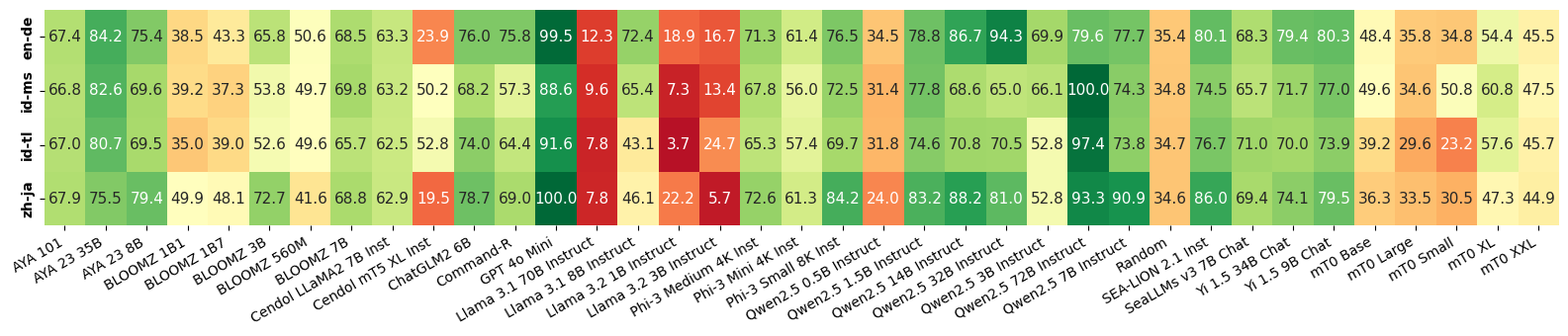}
      \caption{True Cognate}
      \label{fig:true_cognate_understanding}
  \end{subfigure}
  \begin{subfigure}[t]{\linewidth}
      \includegraphics[trim={0, 0, 0, 0}, clip, width=\linewidth]{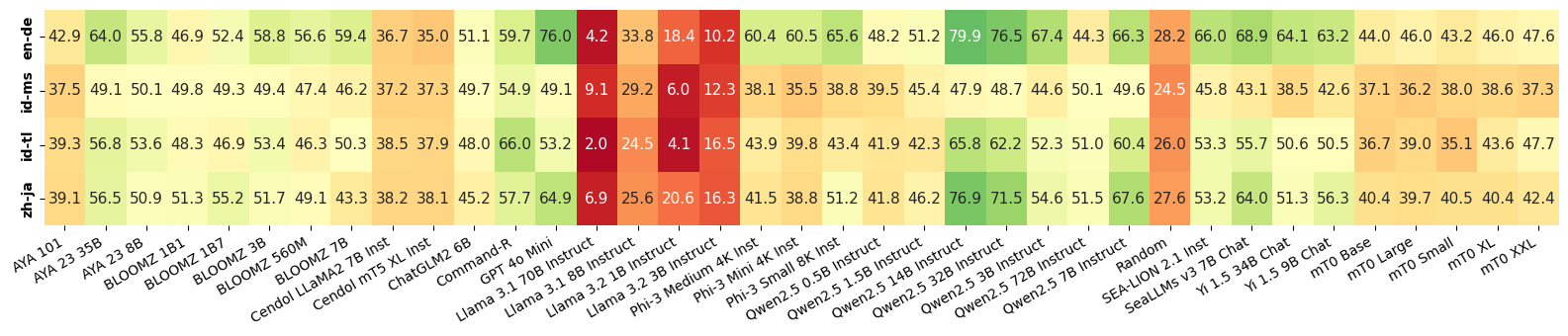}
      \caption{False Friend}
      \label{fig:false_friend_understanding}
  \end{subfigure}
  \caption{Most LLMs understand true cognates, but have limited understanding in regards to false friends in language pairs under study. We report the averaged cognate comprehension scores across the semantic correctness and usage correctness tasks.}
  \label{fig:common_word_understanding}
  \vspace{-8pt}
\end{figure*}

\section{Analysis and Discussion}

We show the stingray plot of the language-and-task aggregated results from our experiment in Figure~\ref{fig:stingray_overall_results}. We observe a clear distinction of LLMs' cognate understanding between true cognate and false friend subsets, and provide further analysis of this behavior in the following section.

\subsection{Do LLMs Understand True Cognates?}

We showcase the breakdown performance per language pair on the true cognate subset in Figure~\ref{fig:true_cognate_understanding}. Although, some smaller-scale LLMs does not perform as good, but larger LLMs tend to yield strong cognate comprehension score. Some LLMs such as Aya-23 (35B), ChatGLM2 (6B), Phi-3 Small, Qwen 2.5, Yi 1.5 (34B), and GPT-4o-mini even achieve almost perfect scores with average cognate comprehension scores $\geq$90\%. This indicates that most LLMs understand the semantics of a true cognate and can incorporate it properly in both languages in the corresponding language pair.

\paragraph{Bias in Cognate Understanding}

\begin{figure*}[!t]
  \centering
  \includegraphics[trim={0, 0, 0, 0}, clip, width=0.95\linewidth]{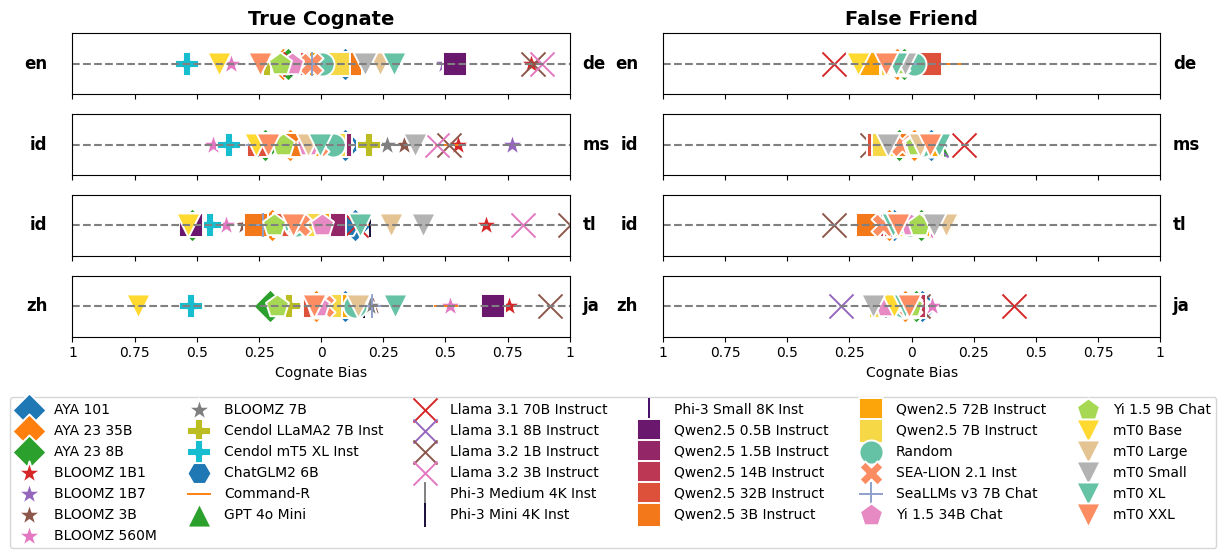}
  \caption{Cognate bias in \textbf{(left)} true cognates and \textbf{(right)} false friends for each language pair under study. We remove the sign of the cognate bias score to avoid confusion.}
  \label{fig:true_cognate_bias}
  \label{fig:false_friend_bias}
  \vspace{-8pt}
\end{figure*}

Although achieving a high cognate comprehension score, some LLMs suffer a high cognate bias. As shown in Figure~\ref{fig:true_cognate_bias}, LLMs such as mT0-XXL and Cendol mT5 XL show strong cognate biases towards relatively higher-resource language in the cognate language pairs including English (in English-German), Indonesian (in Indonesian-Malay and Indonesian-Tagalog), and Chinese (in Chinese-Japanese); while LLMs such as Llama-3.x, BLOOMZ, and MT0 small reflect strong cognate biases towards the other languages. This demonstrates the suitability of \datasetname{} as a testbed for investigating the language selection bias in LLMs~\cite{Nomoto23-chatgpt,nomoto2024masalah}.

\paragraph{Scaling Law and True Cognate Comprehension}
As shown in Figure~\ref{fig:true_cognate_understanding} and Figure~\ref{fig:true_cognate_bias}, we clearly see the impact of the model scale in both cognate bias and cognate comprehension score. For example, BLOOMZ-560M, mT0-small, and Qwen-2.5 0.5B produce low cognate comprehension scores with a high cognate bias, while the larger scale of BLOOMZ, mT0, and Qwen-2.5 have higher cognate comprehension scores with much lower cognate bias. This signifies that the scaling law of language model~\cite{kaplan2020scaling,hoffmann2022training} is advantageous for understanding the sense of true cognates. However, it remains unclear why LLMs of different sizes within the same family exhibit different biases toward certain high-resource and low-resource languages and we leave this exploration for future work.

\subsection{Can LLMs Distinguish False Friend?}

While all LLMs show low cognate bias on the false friend subset as shown in Figure~\ref{fig:false_friend_bias}, most LLMs perform very poorly on the cognate comprehension score in most language pairs. For instance, most LLMs yield comprehension scores that are close to a random baseline as shown in Figure~\ref{fig:false_friend_understanding}. This signifies that most existing LLMs could not even distinguish the sense of false friends across different languages emphasizing an urgent need for a more advanced method on cross-lingual sense disambiguation in multilingual LLMs.

\paragraph{Language Representation Matters}
Despite the quality on the false friend subset is generally low across all LLMs, most LLMs show higher performance on the English-German language pair, including
% This is especially distinct on some LLMs which are English-centric, such as SEALion 2.1 which is based on LLaMA 3 (8B) and Phi model family that achieves cognate comprehension score $\ge$75\%. 
Qwen2.5 (14B and 32B), Yi (34B), Aya 23 (35B), and GPT-4o-mini. This result indicates that most of the existing multilingual LLMs, despite further tuning on other languages, are still English-centric. This observation is consistent with the fact that most existing multilingual LLMs are primarily trained on English data~\cite{Xue2020mT5,muennighoff-etal-2023-crosslingual,chowdhery2023palm,ustun-etal-2024-aya,aryabumi2024aya23openweight}, highlighting the need for enhanced representation of non-English languages. 

\paragraph{Language Similarity Affects False Friend Disambiguation}

We observe that the performance of the Indonesian-Tagalog language pair tends to be higher than the performance of the Indonesian-Malay language pair although the amount of Tagalog data is commonly lower than Malaysian data in the pretraining and supervised fine-tuning (SFT) data. For example, the mC4 corpus~\cite{Xue2020mT5} consists of 0.21\% Malay and only 0. 03\% Tagalog, the pre-training corpus in PaLM~\cite{chowdhery2023palm} consists of 212M Malay tokens and only 175M Tagalog, the Aya dataset~\cite{singh-etal-2024-aya} covers $\sim$4\% of Malay data and <1\% of Filipino (a language closely related to Tagalog) with no Tagalog data. Additionally, although we observe some positive correlation of scaling law in most subsets, we do not observe such a trend in the Indonesian-Malay subset. This signifies that existing LLMs on all scales have difficulty distinguishing false friends between these two languages. 

We hypothesize that this is potentially caused by the high language similarity between Indonesian and Malay. Specifically, Indonesian and Malay fall under the same language family group (Austronesian $\rightarrow$ Malayo-Polynesian $\rightarrow$ Malayic) in both Ethnologue~\cite{kwary2015indomalay,ebenhard2024ethnologue} and Glotollog~\cite{glottolog}. Furthermore, both have great overlap in terms of lexical and grammatical aspects~\cite{kwary2015indomalay,lin2018indomalay,NomotoEtAl_LCC}. Some prior works~\cite{Nomoto23-chatgpt,nomoto2024masalah} have also highlighted that, even a commercial LLM such as ChatGPT~\cite{bang-etal-2023-multitask,wu2023brief,liu2023summary}, still has the problem of differentiating between Malay and Indonesian, and often answers questions in Malay with responses in Indonesian, causing an imbalance of linguistic power, inequality between the two languages, and misrepresentation of the two languages. In this case, we can conclude that disambiguating false friends in language pairs that are highly similar, e.g., Indonesian-Malay, is a significantly more difficult problem compared to a much less similar language pair, e.g., Indonesian-Tagalog.
% , and scaling the data and model capacity might not be sufficient to solve this problem.
%hn: I deleted cole2008indomalay, as I thought it not suitable to cite their paper in this context. I substituted ebenhard2024ethnologue for grimes2000ethnologue,lewis2009ethnologue,ebenhard2021ethnologue.  I thought it's enought to just cite the latest version.

% \subsection{\datasetname{} as a Probe to Language Selection Bias}
% \dummy{\lipsum[5-6]}

\section{Conclusion}

Our work presents a comprehensive evaluation of cross-lingual sense disambiguation in multilingual LLMs. Through the introduction of \datasetname{}~\footnote{Check the \datasetname{}'s dataset card in Appendix~\ref{sec:dataset-card}}, we measure and analyze semantic understanding across languages. By studying false friends and true cognates, we have identified key factors contributing to semantic biases. Our methodology, including the stringray plot and evaluation metrics, i.e., cognate bias and cognate comprehension score, offers a novel approach to understanding cross-lingual sense disambiguation in multilingual LLMs. The generalization of our findings across various language pairs highlights the significance of this work. Our \datasetname{} is not only suitable for measuring the cross-lingual sense disambiguation in LLMs, but also a suitable testbed for investigating language selection bias in multilingual LLMs. We believe that our contributions provide a foundation for further enhancing the cross-lingual capabilities of LLMs, ultimately improving their reliability and performance in diverse linguistic contexts and advancing the development of more inclusive and unbiased multilingual LLMs.
% ensuring accurate and effective multilingual generation.

% Our work has addressed the critical issue of language selection bias in multilingual LLMs, a type of bias that can lead to significant misrepresentation and confusion. By introducing the novel benchmark, \datasetname{}, we have provided a comprehensive evaluation of language selection bias across multiple language pairs, including Indonesian-Malay, Indonesian-Tagalog, Chinese-Japanese, and English-German. Our analysis of false friends, words with similar sounds but different meanings, has shed light on the internal workings of LLMs and potential sources of bias. Through our meticulous task formulation, we have identified factors contributing to language selection bias and proposed insights for mitigating this bias. Our contributions have advanced the development of more inclusive and unbiased language models, ensuring accurate and effective language generation for a diverse range of users.

\section*{Limitation}

\paragraph{Dataset Size} Despite the enormous efforts on annotating with multiple native speakers across different language pairs, due to the limited amount of available false friends and true cognates across different language pairs, our \datasetname{} consists of only around 150-200 samples per language pairs. To cater to this limitation, we try to increase the task, allowing probing of multilingual LLMs with bigger sample sizes. We leave further exploration on how to increase the amount of data of false friend and true cognate to future work. 

\paragraph{Benchmark Coverage} Due to the difficulty in finding annotators, our \datasetname{} only covers four language pairs, i.e., English-German, Indonesian-Malay, Indonesian-Tagalog, and Chinese-Japanese. There are many other potential language pairs that can be covered in the benchmark, such as Sloven-Croatian, Spanish-Portuguese, etc. We expect future work to extend the generalization of our benchmark and findings to other language pairs.

\section*{Ethics Statement}

This work introduces a novel benchmark for cross-lingual sense disambiguation and evaluation in multilingual large language models (LLMs), aiming to uncover biases and limitations in their semantic understanding across languages. Throughout the development of this benchmark, several ethical considerations were taken into account.

\paragraph{Inclusivity and Fairness} The primary motivation of our work is to highlight and address the biases present in multilingual LLMs, particularly toward high-resource languages. We recognize that current language technologies often underperform speakers of low-resource languages, which could reinforce language hierarchies and contribute to the marginalization of these linguistic communities. By incorporating language pairs such as Indonesian-Malay and Indonesian-Tagalog, we strive to promote inclusivity and fairness in the evaluation of LLMs and advocate for broader linguistic diversity in NLP research.

\paragraph{Bias and Misrepresentation} One of the key goals of our research is to identify bias in cross-lingual semantic disambiguation, especially concerning the handling of false friends and true cognates. We understand that biases in LLMs can result in misrepresentation of user intent and can have far-reaching consequences when applied in real-world scenarios. Our benchmark seeks to pinpoint these issues, providing tools for researchers and practitioners to mitigate such biases and ensure that LLMs produce more accurate and fair multilingual outputs.

\paragraph{Data Annotation} The data used in our benchmark was carefully curated and annotated by native speakers of the respective languages to ensure linguistic accuracy and cultural sensitivity. We made every effort to fairly compensate our annotators and ensure that their contributions were recognized and valued. Additionally, we acknowledge the limitations of our dataset size and coverage and encourage further efforts to expand and diversify the benchmark in future work.

\paragraph{Privacy and Security} Our dataset does not include any personally identifiable information or sensitive data. The false friends and true cognates were collected from publicly available resources, and no private or proprietary data were used in this research. We ensured that all data collection and usage adhered to ethical guidelines and standards in the field of natural language processing. 

By addressing these ethical considerations, we aim to foster more responsible and equitable multilingual LLMs, contributing to the advancement of fair and inclusive language technologies.

\bibliography{custom}

\newpage

\appendix

\section{Annotation Guideline}
\label{sec:annotation-guideline}

\subsection{Annotation Objective}

The goal of this annotation task is to create a dataset that distinguishes between false friends and true cognates across various language pairs. Annotators will work with native speakers to identify and categorize common words, construct sentences, and translate them to ensure accurate representation.

\subsection{Word Selection and Criteria}

\textbf{False Friends}: Words with the same spelling or characters but different meanings in the respective languages. \textbf{True Cognates}: Words with the same spelling, characters, and meanings in both languages. For collecting the common words, annotators incorporate the following sources:
\begin{itemize}
    \item Wiktionary's lists of false friends~\footnote{\url{https://en.wiktionary.org/wiki/Category:False_cognates_and_false_friends}}
    \item Kamus Komunikatif Nusantara: Indonesia-Malaysia, Malaysia-Indonesia~\citep{MohdSharifudinAlMudra2015}
    \item Kamus Kata: Bahasa Melayu Malaysia-Bahasa Indonesia~\citep{Rusdi2016}
\end{itemize}

\subsection{Annotation Process}

\paragraph{Annotation Flow}
\begin{itemize}
    \item \textbf{Translation and Replacement} For false friends, given a false friend word, an annotator will make the correct sentence in their language and translate it into English. The annotator of the other language in the pair will then translate the English translation into their native language. The target false friend word will be replaced with a different word that conveys the intended meaning. This will result in a semantically odd sentence.
    \item \textbf{Cognate Agreement and Translation}: For true cognates, both annotators will first agree on an English sentence. The English sentence will then be translated into their respective native languages to construct the true cognate sentence pair.
\end{itemize}

\paragraph{Allowed Variations}
\begin{itemize}
    \item English-German: Words differ in capitalization with a maximum of one edit distance.
    \item Chinese-Japanese: Words with different characters developed from the same origin.
    \item Indonesian-Malay: Exact match words.
    \item Indonesian-Tagalog: Words with maximum of one edit distance.
\end{itemize}

\subsection{Dataset Entries}
\begin{itemize}
    \item \textbf{False Friends}: Each false friend will typically have two entries --- one for the correct usage and one for the incorrect usage.
    \item \textbf{True Cognates}: Each true cognate will have one entry, as both native language sentences translate correctly.
\end{itemize}

\subsection{Examples}

\textbf{False Friend (Indonesian-Tagalog):}
\begin{itemize}
    \item Indonesian sentence: Pagi ini saya begitu senang
    \item English translation: This morning, I feel so happy
    \item Tagalog translation: Ako ay masaya ngayong umaga
    \item Replacement: Ako ay masaya ngayong pagi ("Today's stingray, I feel happy")
\end{itemize}

\textbf{True Cognate (Indonesian-Malay)}
\begin{itemize}
    \item English sentence: "That apple has many caterpillars."
    \item Indonesian sentence: Apel itu banyak ulat
    \item Malay sentence: Epal itu ada banyak ulat.
\end{itemize}

\subsection{Additional Guidance for Annotators}
\begin{itemize}
    \item Ensure a clear understanding of the word's meaning and context.
    \item Construct sentences that are natural and grammatically correct in your native language.
    \item Pay attention to the nuances and potential variations in word usage.
    \item For false friends, aim for a semantically odd translation to highlight the semantic differences between the two sentences.
    \item Collaborate effectively with your partner annotator to ensure accurate translations and representations.
\end{itemize}

\section{Dataset Card}
\label{sec:dataset-card}
\textbf{Dataset Name}: StingrayBench

\subsection{Dataset Description}

\paragraph{Overview} StingrayBench is a dataset designed to evaluate models' understanding of semantic appropriateness and cognate word usage across multiple language pairs. The dataset focuses on false friends and true cognates, which are words with similar spellings or characters but different meanings or additional meanings in different languages.

\paragraph{Language Pairs} The dataset covers the following language pairs:

\begin{itemize}
    \item English-German (\textsc{en-de})
    \item Chinese-Japanese (\textsc{zh-ja})
    \item Indonesian-Malay (\textsc{id-ms})
    \item Indonesian-Tagalog (\textsc{id-tl})
\end{itemize}

\subsection{Dataset Construction}

Native speakers of the respective languages were involved in constructing the dataset. They listed common words between language pairs and their meanings, consulting resources on false friends. The words were then categorized as false friends or true cognates.For each word, annotators created sentences in their native language and provided English translations. The sentences were designed to showcase the correct and incorrect usage of the target words. In the case of false friends, the sentences were manipulated to produce semantically odd translations.

\subsection{Dataset Statistics}

The dataset contains a total of 705 entries, including: 259 true cognate entries and 446 false friend entries. The distribution of entries across language pairs is as follows:

\begin{itemize}
    \item \textsc{en-de}: 196 entries (98 true cognates, 98 false friends)
    \item \textsc{zh-ja}: 165 entries (51 true cognates, 114 false friends)
    \item \textsc{id-ms}: 186 entries (52 true cognates, 134 false friends)
    \item \textsc{id-tl}: 158 entries (58 true cognates, 100 false friends)
\end{itemize}

\subsection{Task Formulation}

\paragraph{Semantic Appropriateness}

In this task, models are prompted to determine which sentence is more semantically appropriate. The prompt includes two sentences from the language pair and a third option indicating that both sentences are appropriate. This task aims to test the model's comprehension and understanding of the semantic nuances between the language pairs.

\paragraph{Usage Correction}

The usage correction task focuses on the correct usage of specific cognate words. Models are prompted with a sentence containing a cognate word and asked to determine if the word's usage is correct. This task provides a more targeted evaluation of the model's ability to handle cognate words accurately.

\subsection{Example Prompts and Completions}

\paragraph{Semantic Appropriateness}.

\noindent\fbox{%
    \parbox{\linewidth}{%
        \textbf{Prompt:} \newline
        Which sentence is more semantically appropriate? \newline
        A. "Ich habe einen Arm." (German) \newline
        B. "I have an Arm." (English) \newline
        C. "Both sentences are appropriate."
        \newline
        \newline
        \textbf{Target Completion:} "C. Both sentences are appropriate."
    }
}
\paragraph{Usage Correction}.

\noindent\fbox{%
    \parbox{\linewidth}{%
        \textbf{Prompt:} \newline
        Is the usage of "pagi" in this sentence correct? "Ako ay masaya ngayong pagi." (Tagalog)
        \newline
        \newline
        \textbf{Target Completion:} \newline
        "No, the usage of 'pagi' is incorrect. 'Pagi' means 'stingray' in Tagalog, and the sentence should use 'umaga' for 'morning'."
    }
}

\subsection{Dataset Licensing Information}

To promote accessibility, encourage collaboration, and facilitate knowledge sharing, \datasetname{} will be made available to the public under the Creative Commons Attribution-ShareAlike 4.0 International (CC-BY-SA 4.0) license. This license ensures that the dataset is accessible and can be utilized by a wide range of individuals and organizations including for commercial users.

\section{Additional Results}
\label{sec:additional-results}

\subsection{Stingray Plot}

\paragraph{Overall}

\begin{figure*}[h]
  \centering
  \includegraphics[trim={0, 0, 0, 0}, clip, width=0.95\linewidth]{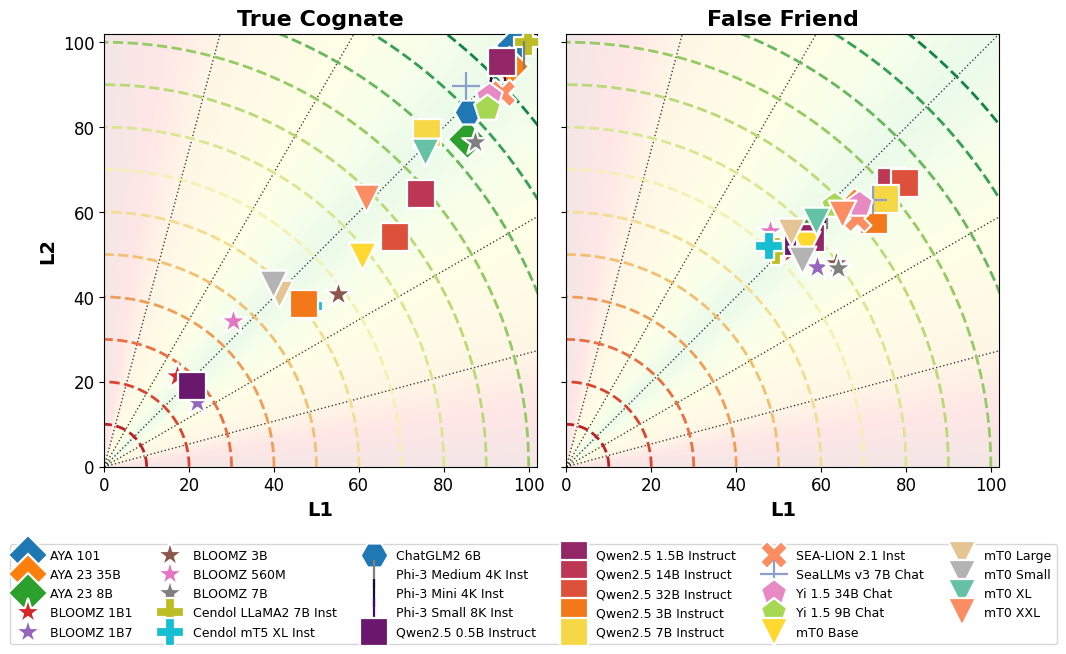}
  \caption{Stingray plot showcasing overall cognate understanding of \textbf{(left)} true cognates and \textbf{(right)} false friends in \textbf{usage correctness}.}
  \label{fig:stingray_usage_correctness_overall}
  \vspace{-8pt}
\end{figure*}

\begin{figure*}[h]
  \centering
  \includegraphics[trim={0, 0, 0, 0}, clip, width=0.95\linewidth]{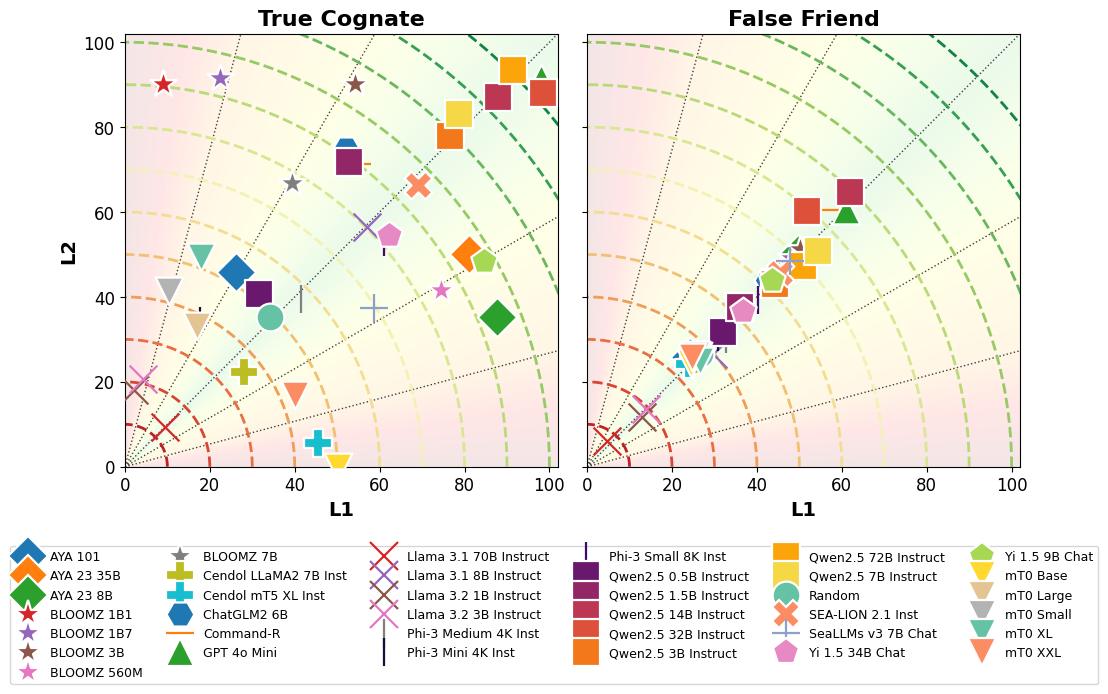}
  \caption{Stingray plot showcasing overall cognate understanding of \textbf{(left)} true cognates and \textbf{(right)} false friends in \textbf{usage correctness}.}
  \label{fig:stingray_semantic_correctness_overall}
  \vspace{-8pt}
\end{figure*}

Figure~\ref{fig:stingray_usage_correctness_overall} and \ref{fig:stingray_semantic_correctness_overall} respectively show overall cognate understanding of true cognates and false friends in the usage correctness task and the semantic correctness task.

\paragraph{Per language pair}

\begin{figure*}[h]
  \centering
  \includegraphics[trim={0, 0, 0, 0}, clip, width=0.7\linewidth]{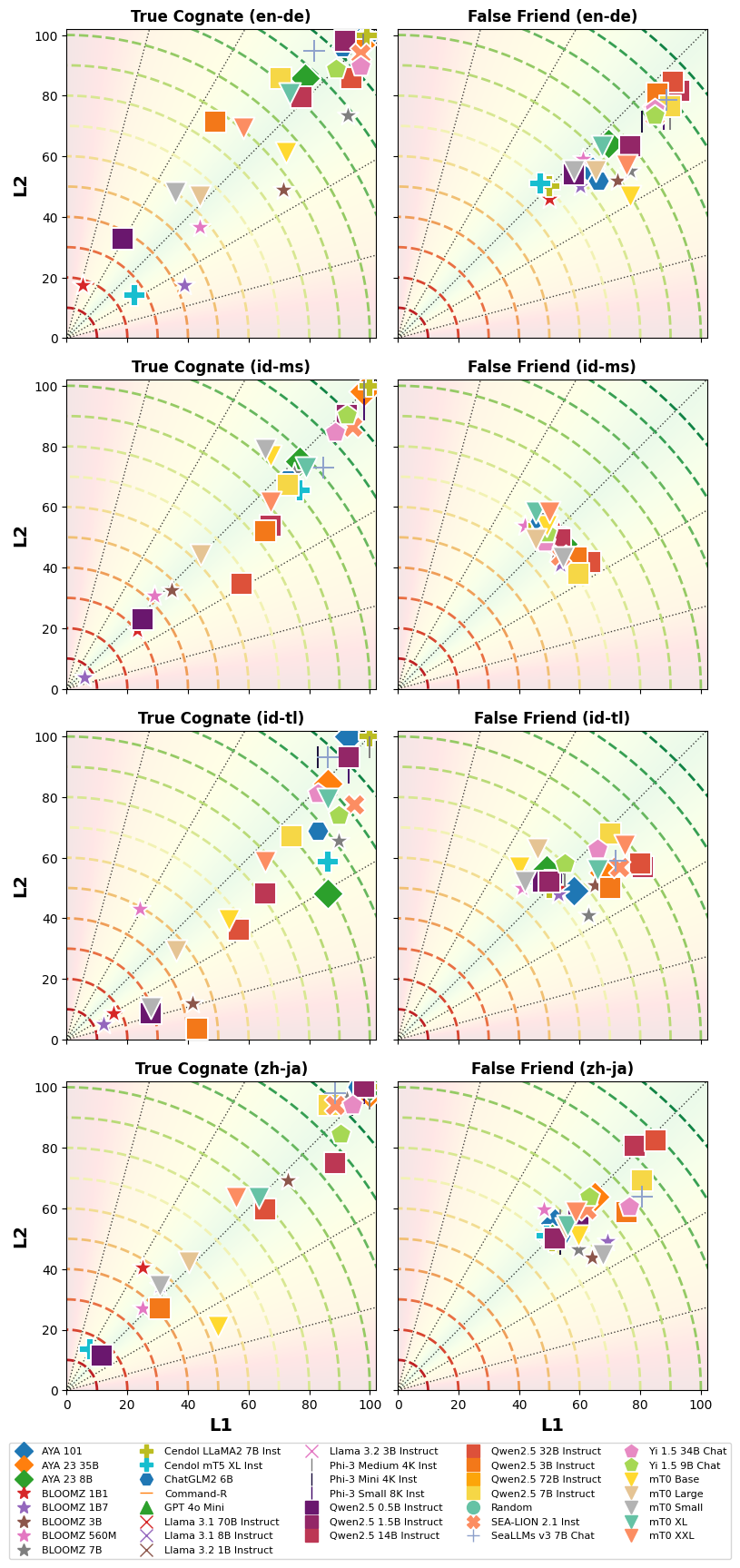}
  \caption{Stingray plot showcasing cognate understanding of \textbf{(left)} true cognates and \textbf{(right)} false friends in \textbf{usage correctness} per language pair under study.}
  \label{fig:stingray_usage_correctness_per_lang_pair}
  \vspace{-8pt}
\end{figure*}

\begin{figure*}[h]
  \centering
  \includegraphics[trim={0, 0, 0, 0}, clip, width=0.7\linewidth]{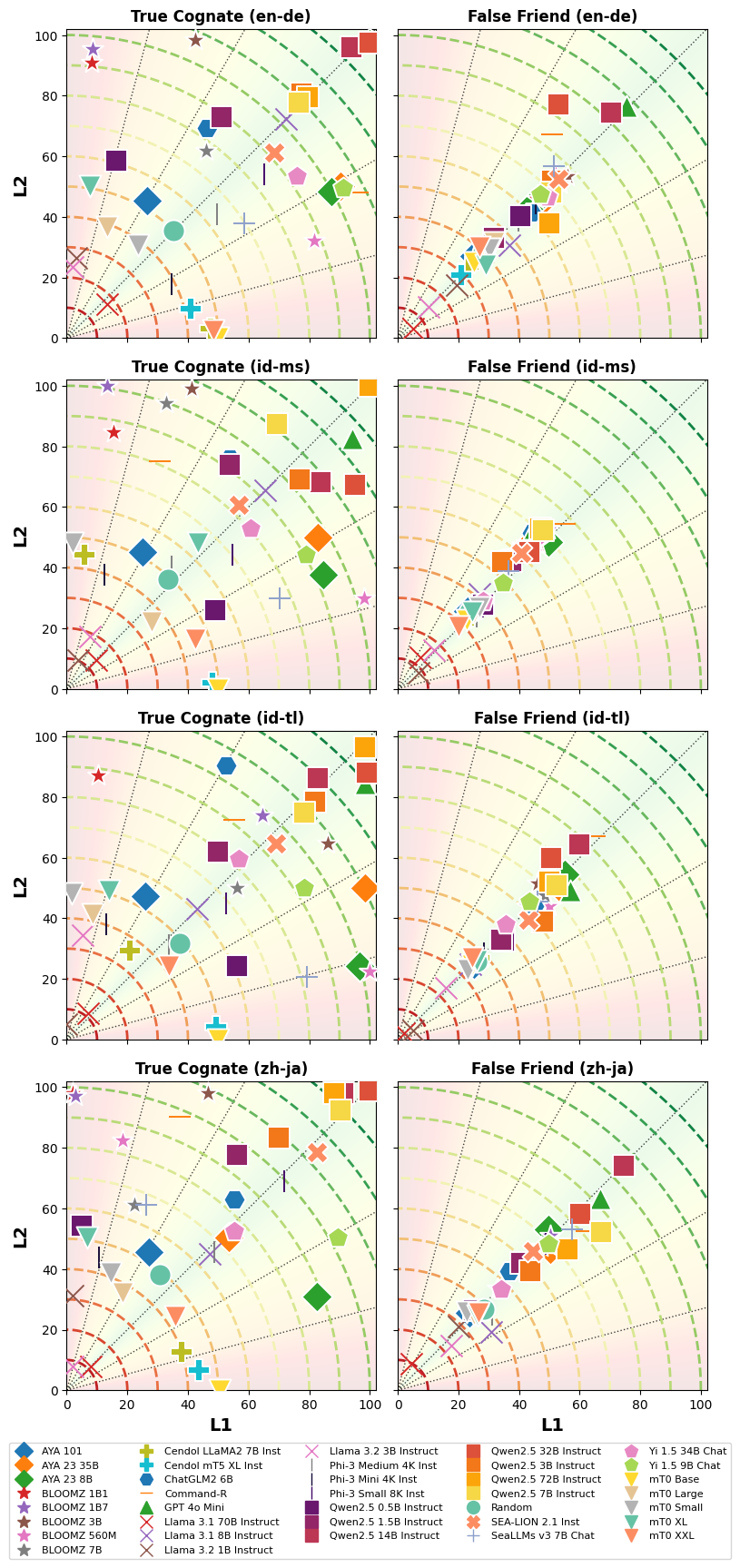}
  \caption{Stingray plot showcasing cognate understanding of \textbf{(left)} true cognates and \textbf{(right)} false friends in \textbf{semantic correctness} per language pair under study.}
  \label{fig:stingray_semantic_correctness_per_lang_pair}
  \vspace{-8pt}
\end{figure*}

Figure~\ref{fig:stingray_usage_correctness_per_lang_pair} and \ref{fig:stingray_semantic_correctness_per_lang_pair} respectively show cognate understanding of true cognates and false friends in the usage correctness task and the semantic correctness task for all language pairs under study.

\subsection{Cognate Comprehension}

\paragraph{Usage Correctness}

\begin{figure*}[h]
  \centering
  \begin{subfigure}[h]{\linewidth}
      \includegraphics[trim={0, 6.6em, 0, 0}, clip, width=\linewidth]{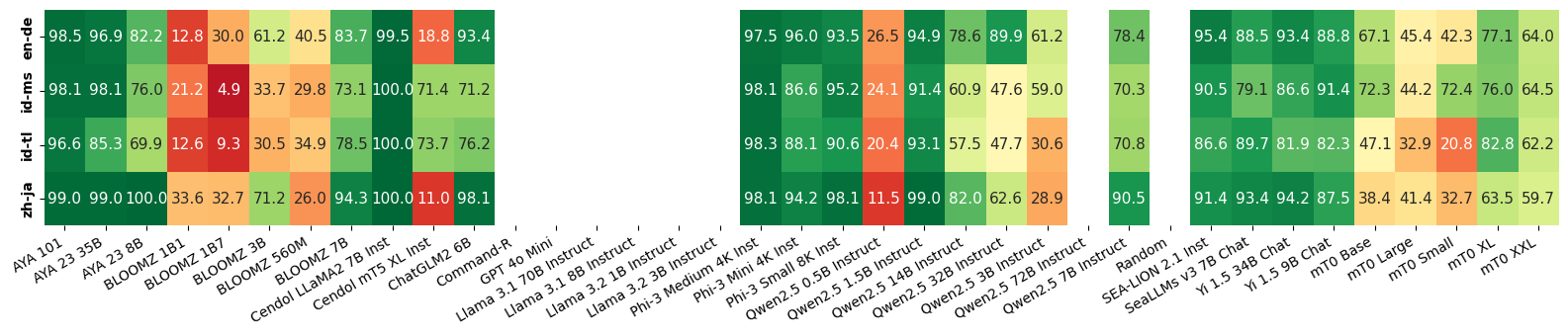}
      \caption{True Cognate}
      \label{fig:true_cognate_understanding_usage_correctness}
  \end{subfigure}
  \begin{subfigure}[h]{\linewidth}
      \includegraphics[trim={0, 0, 0, 0}, clip, width=\linewidth]{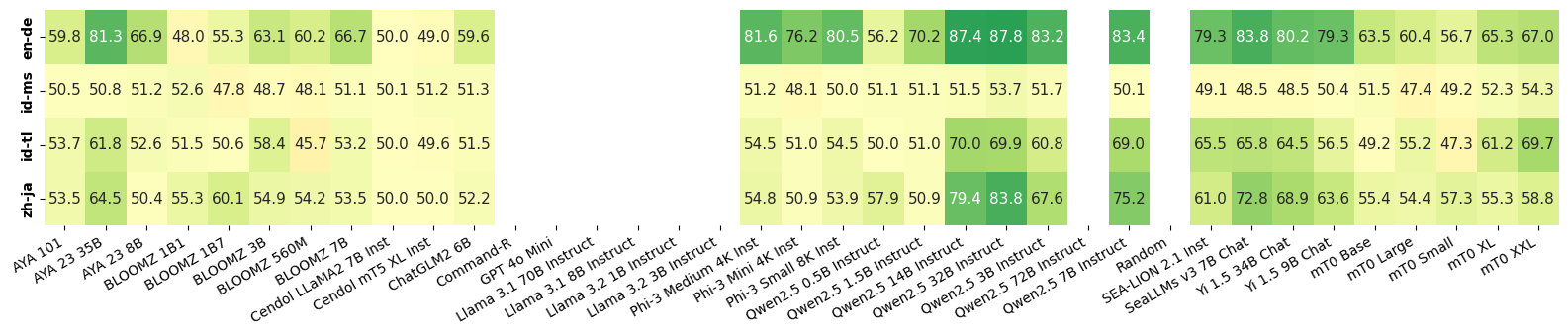}
      \caption{False Friend}
      \label{fig:false_friend_understanding_usage_correctness}
  \end{subfigure}
  \caption{Cognate comprehension of LLMs in \textbf{usage correctness} task.}
  \label{fig:common_word_understanding_usage_correctness}
  \vspace{-8pt}
\end{figure*}

Figure~\ref{fig:true_cognate_understanding_usage_correctness} and \ref{fig:false_friend_understanding_usage_correctness} respectively show cognate comprehension of true cognates and false friends in the usage correctness task.

\paragraph{Semantic Correctness}

\begin{figure*}[h]
  \centering
  \begin{subfigure}[h]{\linewidth}
      \includegraphics[trim={0, 6.6em, 0, 0}, clip, width=\linewidth]{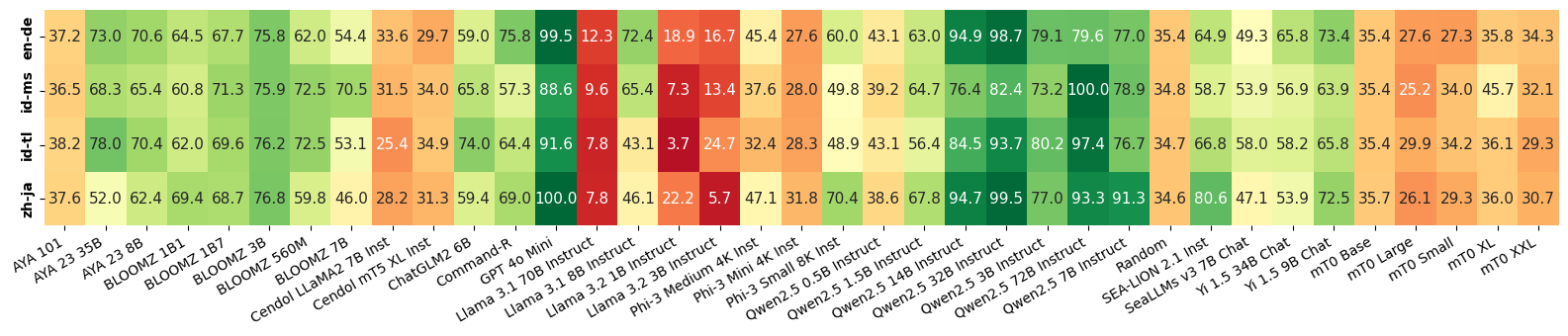}
      \caption{True Cognate}
      \label{fig:true_cognate_understanding_semantic_correctness}
  \end{subfigure}
  \begin{subfigure}[h]{\linewidth}
      \includegraphics[trim={0, 0, 0, 0}, clip, width=\linewidth]{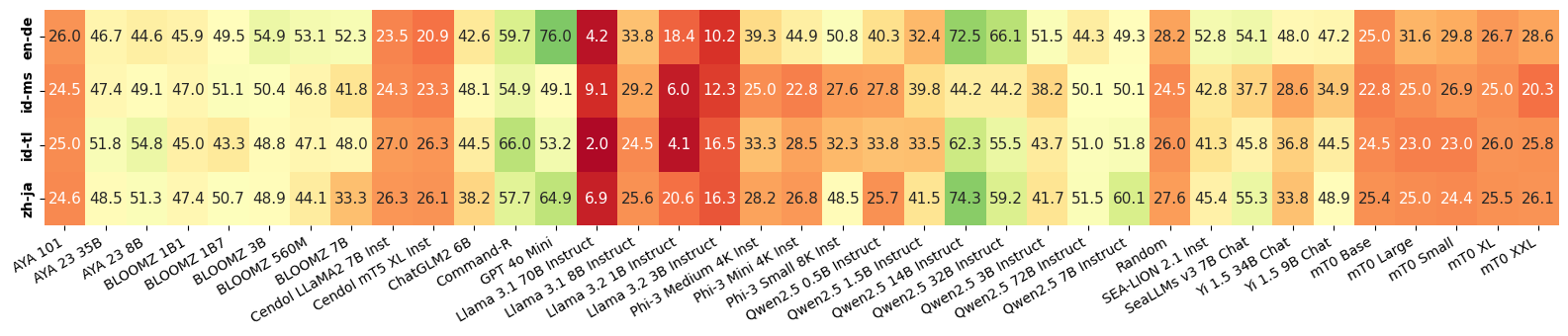}
      \caption{False Friend}
      \label{fig:false_friend_understanding_semantic_correctness}
  \end{subfigure}
  \caption{Cognate comprehension of LLMs in \textbf{semantic correctness} task.}
  \label{fig:common_word_understanding_semantic_correctness}
  \vspace{-8pt}
\end{figure*}

Figure~\ref{fig:true_cognate_understanding_semantic_correctness} and \ref{fig:false_friend_understanding_semantic_correctness} respectively show cognate comprehension of true cognates and false friends in the semantic correctness task.

\subsection{Cognate Bias}

\paragraph{Usage Correctness}

\begin{figure*}[h]
  \centering
  \includegraphics[trim={0, 0, 0, 0}, clip, width=\linewidth]{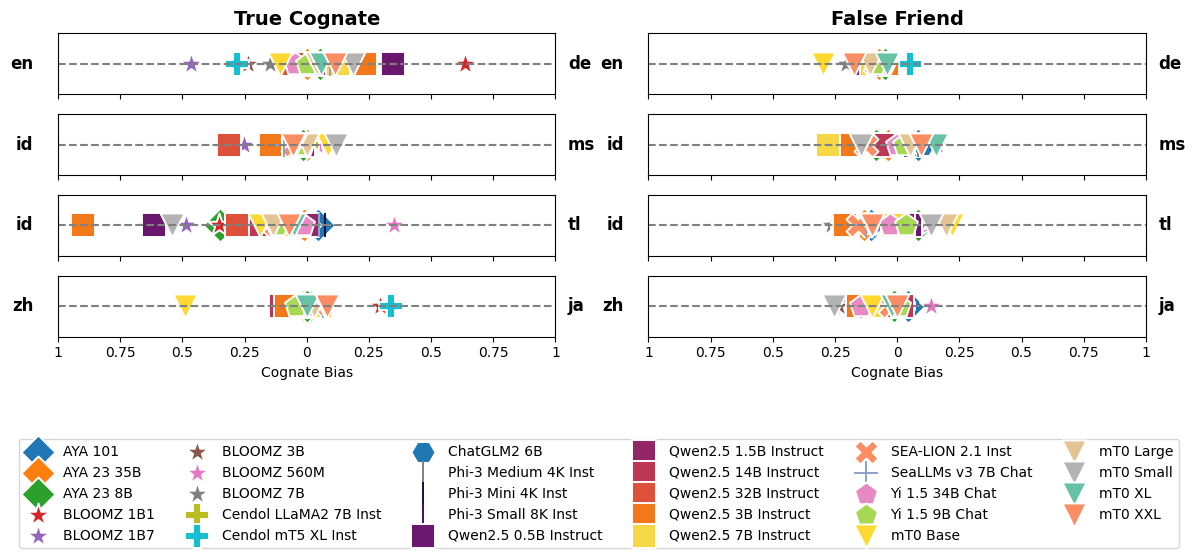}
  \caption{Cognate bias on \textbf{(left)} true cognates and \textbf{(right)} false friends in \textbf{usage correctness} per language pair under study.}
  \label{fig:stingray_bias_usage_correctness}
  \vspace{-8pt}
\end{figure*}

Figure~\ref{fig:stingray_bias_usage_correctness} shows cognate bias of true cognates and false friends in the usage correctness task.

\paragraph{Semantic Correctness}

\begin{figure*}[h]
  \centering
  \includegraphics[trim={0, 0, 0, 0}, clip, width=\linewidth]{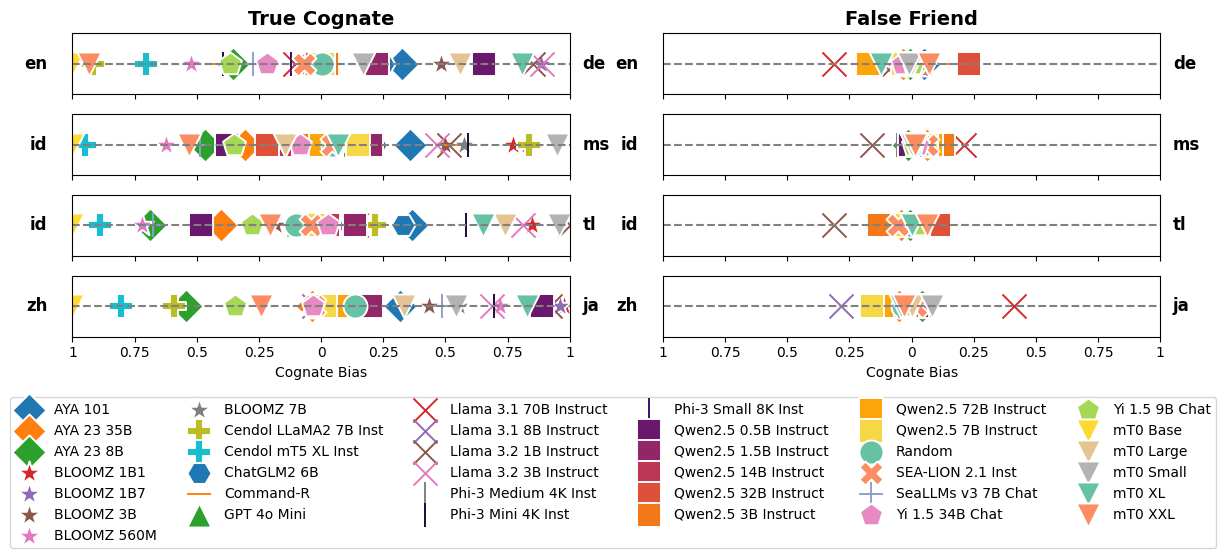}
  \caption{Cognate bias on \textbf{(left)} true cognates and \textbf{(right)} false friends in \textbf{semantic correctness} per language pair under study.}
  \label{fig:stingray_bias_semantic_correctness}
  \vspace{-8pt}
\end{figure*}

Figure~\ref{fig:stingray_bias_semantic_correctness} shows cognate bias of true cognates and false friends in the semantic correctness task.

\end{document}